\documentclass{article}

\PassOptionsToPackage{numbers}{natbib}
\usepackage[final]{neurips_2024}

\usepackage{graphicx}

\usepackage[utf8]{inputenc} % allow utf-8 input
\usepackage[T1]{fontenc}    % use 8-bit T1 fonts
\usepackage{hyperref}       % hyperlinks
\usepackage{url}            % simple URL typesetting
\usepackage{booktabs}       % professional-quality tables
\usepackage{amsfonts}       % blackboard math symbols
\usepackage{nicefrac}       % compact symbols for 1/2, etc.
\usepackage{microtype}      % microtypography
\usepackage{xcolor}         % colors

\usepackage{comment}
\usepackage{caption}
\usepackage{tablefootnote}

\title{Dual-Model Distillation for Efficient Action Classification with Hybrid Edge-Cloud Solution}

\author{
  Timothy Wei~\thanks{Work done during internship at the University of California, Santa Cruz}\\
  Saratoga High School\\
  \texttt{timswei@gmail.com} \\
  \and
  \textbf{Hsien Xin Peng~\footnotemark[1]}\\
  Ridge High School\\
  \texttt{maxpeng123678@gmail.com} \\
  \and
  \textbf{Elaine Xu} \\
  The Harker School\\
  \texttt{elainexuyz@gmail.com} \\
  \and
  \textbf{Bryan Zhao~\footnotemark[1]}\\
  Saratoga High School\\
  \texttt{bryanzhao08@gmail.com} \\
  \and
  \textbf{Lei Ding} \\
  University of California Santa Cruz\\
  \texttt{lding25@ucsc.edu} \\
  \and  
  \textbf{Diji Yang~\thanks{Corresponding to: dyang39@ucsc.edu}}\\
  University of California Santa Cruz\\
  \texttt{dyang39@ucsc.edu} \\
}

\begin{document}

\maketitle

\begin{abstract}

As Artificial Intelligence models, such as Large Video-Language models (VLMs), grow in size, their deployment in real-world applications becomes increasingly challenging due to hardware limitations and computational costs. To address this, we design a hybrid edge-cloud solution that leverages the efficiency of smaller models for local processing while deferring to larger, more accurate cloud-based models when necessary. Specifically, we propose a novel unsupervised data generation method, Dual-Model Distillation (DMD), to train a lightweight switcher model that can predict when the edge model’s output is uncertain and selectively offload inference to the large model in the cloud. Experimental results on the action classification task show that our framework not only requires less computational overhead, but also improves accuracy compared to using a large model alone. Our framework provides a scalable and adaptable solution for action classification in resource-constrained environments, with potential applications beyond healthcare. Noteworthy, while DMD-generated data is used for optimizing performance and resource usage in our pipeline, we expect the concept of DMD to further support future research on knowledge alignment across multiple models.
\end{abstract}

\section{Introduction}
% large model is good, but too large
Recent advances in Artificial Intelligence (AI) have shown that increasing model size leads to substantial performance gains across various domains, from language models~\citep{dubey2024llama, jiang2023mistral} to video-language models~\citep{lin2023video, wunext}. Large-scale models, with billions of parameters, consistently achieve state-of-the-art results on various benchmarks~\citep{zhao2023survey}. However, these models can be challenging to deploy when translating benchmark performance into real-world applications~\citep{minaee2024large} due to their high runtime costs.

% application field
One critical area where this limitation becomes evident is in action classification for elderly care~\citep{sawik2023robots}. A task such as predicting falls through non-intrusive monitoring~\citep{maldonado2019fallen, elwaly2024new} is a critical yet complex task that benefits from large-scale video-language models~\citep{elwaly2024new}. Despite their high accuracy, these oversized models are not feasible for deployment on common elderly care appliances due to hardware constraints~\citep{chen2019deep}. 
To address this, we propose a hybrid edge-cloud solution where small models run locally on edge devices while larger models deployed in the cloud are accessed only when necessary.
% technical background
However, the core technical challenge here remains: how to effectively align the knowledge and capabilities of the small and large models to ensure seamless collaboration between them. 
While Model or Knowledge Distillation techniques~\citep{hinton2015distilling, gou2021knowledge} have been extensively explored, their primary goal is to compress the knowledge of large models into smaller ones. Yet, this process often falls short in collaborative systems where both small and large models have unique strengths. The smaller model, after distillation, may not fully exploit its distinct capabilities or optimally collaborate with its larger counterpart~\citep{narayan2022predicting}.
% One of the mainstream efforts is known as Model/Knowledge Distillation, where a large model serves as a teacher to teach the smaller student model~\cite{hinton2015distilling} \cite{gou2021knowledge}. However, smaller models are limited by size and can only asymptote but not reach the performance of larger models. More crucially, in collaborative systems where edge and cloud models (large and small) must work together, simply distilling the larger model’s knowledge into the smaller one may not optimize performance~\citep{narayan2022predicting}. Each model may have distinct areas of expertise, and replicating the large model’s capabilities in the small model does not guarantee better collaboration between them. In these scenarios, relying on knowledge distillation could even limit the synergy between the models, as the smaller model may miss opportunities to independently leverage its unique strengths.

% We propose
To overcome these limitations, we introduce Dual-Model Distillation (DMD), a novel approach to enhance cooperation between small and large models. 
Unlike traditional distillation, which focuses on unidirectional knowledge transfer, DMD treats both models as teacher models to co-teach and generate synthetic data to train a student model (i.e., the switcher model).
% Learned from DMD-generated data, at inference time, the switcher model decides the best timing to invoke the cloud-deployed large model. 
% The DMD process begins by collecting inference data from teacher models separately. Since the switcher model only requires ``agree'' or ``disagree'' labels, no additional human labeling effort or domain-specific datasets are needed. Instead, the agreement verdict between zero-shot inference outputs of the small and large models is enough to train the switcher model. 
The DMD process eliminates the need for additional human-labeled data by leveraging the agreement or disagreement between the two teacher models’ zero-shot inferences to create training data.
Through instruction-supervised finetuning~\citep{chung2024scaling}, the switcher model learns to make optimal decisions about when to invoke the larger model.
% Results
Experiments conducted on the dataset used in Fall Detection - Eldercare Robot datase~\cite{Elwaly2023dataset} demonstrate the effectiveness of our approach. Surprisingly, our pipeline not only demands less computational overhead than just using the large model, resulting in 39.5\% cost savings, but also outperforms the isolated large model by 4.6\% in terms of F1 score. Furthermore, we anticipate that the DMD concept will provide substantial opportunities for future research on knowledge integration across multiple deep learning models.

\section{Methodology}
\label{sec:approach}
%%%%% Present idea and the system overview
% i.e., what does the pipeline look like

\begin{figure}
    \centering
    \includegraphics[scale=0.9, width=0.90\linewidth]{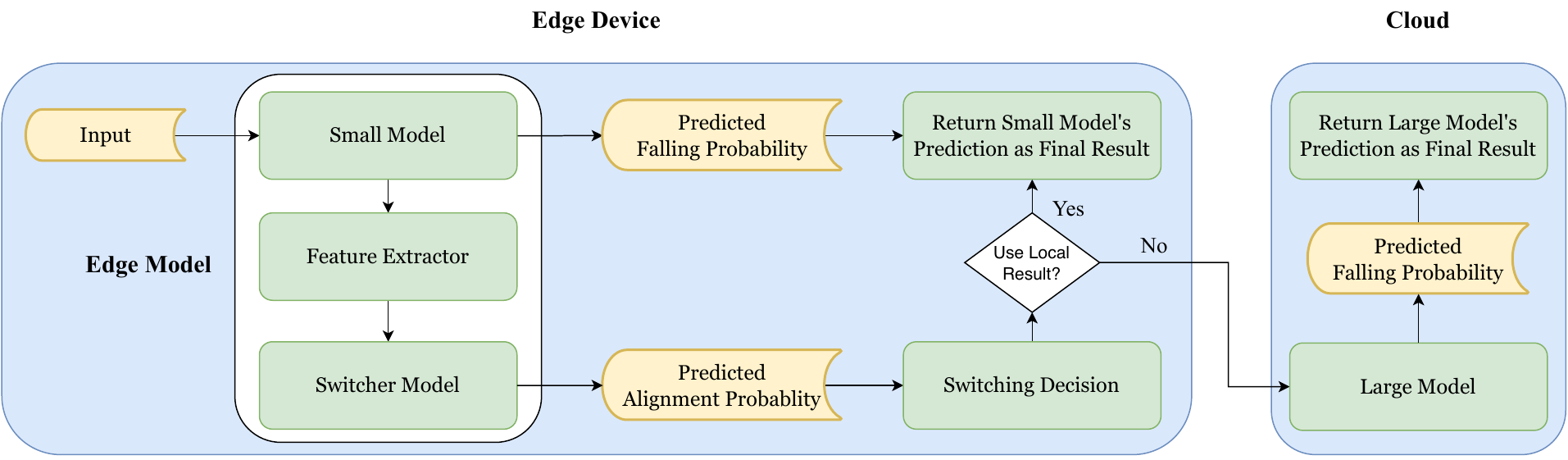}
    \caption{Overview of our hybrid edge-cloud solution}
    \label{fig:fig1}
\end{figure}

% As shown in Figure~\ref{fig:fig1}, our system involves three models: the small, switcher and large models. The small model's role is to provide a preliminary output as well as its reasoning. Then, the switcher model considers the small model's reasoning and decides whether to keep the small model's output or to call the large model and take its output. By only calling the large model when necessary, we save computational cost while optimizing accuracy. Below, we provide the details of each key component of our framework.

% As shown in Figure~\ref{fig:fig1}, the small model takes in the image input. Then, we extract the small model's ``reasoning'' from hidden layers within the model and convert them into the form of a tensor. We input this tensor into our switcher model, which processes it and outputs a probability indicating how likely it thinks that the large model will agree with the small model's prediction. We then conducted the analysis to establish an optimal threshold for this probability to obtain the most accurate overall results. Below the threshold probability, the small model's output is taken. Otherwise, the large model is called, and its output is taken over that of the small model.

% motivation of 1, reasoning based on hidden features 2, switching decision based on budget control/uncertainty percentile analysis
As shown in Figure~\ref{fig:fig1}, our hybrid edge-cloud solution introduces a companion switcher model using the hidden features of the input data extracted from the small model, which serves as the foundation of its reasoning capacity. With the help of the switcher model, our edge model can not only predict the falling probability but also accurately estimate the probability that the large model's result aligns with its own. By incorporating this switcher model, the edge model deliberates when to invoke the cloud model, thereby keeping system expenses under a pre-defined budget. In our case, a threshold of alignment probability, which yields optimal performance in predicting the falling probability during training, determines when to offload data to the large model during inference time. Regarding more complex decision-making scenarios, we expect that a more sophisticated utility-based threshold design \cite{wang2011utilizing} can also be applied in the future.

\paragraph{Dual-Model Distillation}
To give the switcher model an understanding of the capabilities of both models, we propose a novel training data generation method. First, we pass all the raw data through the small model and extract its hidden representation together with its prediction. Then, we trigger another run-through using the large model to generate the ground truth of alignment, or agreement, between the predictions of the small and large models. It is important to underscore the generalizability of this method, as for any of two deep learning models with the same I/O format, the DMD process can generate the above-described data from any raw data with only the required initial input and final output (which is provided in most datasets). During the subsequent instruction finetuning, the switcher model learns the dual-model distilled knowledge. At inference time, using its grasp of both models' capability, the switcher is able to decide when to utilize the power of the large model.

\section{Experiments}
%%%%% Introduce the task and the data  
We conducted the experiments on the Fall Detection - Eldercare Robot~\citep{Elwaly2023dataset} dataset, with more details provided in Appendix~\ref{appendix:datasets}.
% \paragraph{Implementation Details} 
We used a Vision-and-Language Transformer (ViLT) \cite{kim2021vilt} as the small model, chosen for its computational efficiency and relatively small size (less than 500 MB). For the large model, we utilized LLaVA-NeXT-Video-7B~\citep{li2024llava}, which is a well-performing open-source video-language model. Following the DMD process described in Section~\ref{sec:approach}, we generated data where the inputs consist of the last hidden layer of the small model for each image input in the raw data, and the outputs indicate whether the small model and large model agree on their falling predictions. To minimize the computing on edge devices, we design a lightweight switcher model implemented as a Multi-Layer Perceptron (MLP) as an ad-hoc layer on top of the small model. The switcher model is trained on the DMD-generated data for the intelligent model switching. More implementation details are available in Appendix~\ref{appendix:implementation_details}.

\subsection{Results}
%%% Considering change the table to large table
For performance comparison, we conducted experiments and included the results of three baselines. As reported in Table~\ref{tab:main_result}, reliance solely on the ViLT (Small model only) resulted in an F1 score of 58.2\%, while LLaVA-NeXT-Video (Large model only) got 87.5\%. In addition, we reproduced a popular uncertainty-based switcher model approach~\citep{narayan2022predicting} on our task. Specifically, we extracted the falling probability in the final output layer of ViLT and used that as the metric to determine whether to defer judgment to the larger model. This baseline got 76.1\% when offloading 60\% computing to the large model. Our system achieves 92.1\%, which is 33.9\%, 4.6\%, and 16.0\% higher than the traditional approach, large model-only approach and the uncertainty-based approach respectively. 

\paragraph{Performance Analysis}
%% To be updated
To gain a deeper understanding of the performance, we manually adjust the threshold for offloading decisions to large models as described in~\ref{appendix:threshold}. As shown in Figure~\ref{fig:F1_data_percentage_plot_on_testset}, our system consistently outperforms the traditional uncertainty-based system regardless of the offload threshold. Furthermore, it suggests that solely calling the large model may be \textit{less optimal} than calling it only when the switcher model indicates to do so, as the switcher model knows better about both models' capabilities from the DMD process. Therefore, we conclude that the switcher model offers a low-cost, lightweight solution to increase fall detection accuracy while optimizing efficiency.

% TODO: this table should be changed 

\begin{comment}    
\begin{figure}[htbp]
    \centering
    \begin{minipage}{0.45\linewidth}
        \centering
        \includegraphics[width=\linewidth]{refined-graph.png}
        \caption{Combined and baseline model accuracy against the percentage of data deferred to the large model}
        \label{fig:F1_data_percentage_plot_on_testset}
    \end{minipage}%
    \hfill
    \begin{minipage}{0.45\linewidth}
        \centering
        \begin{tabular}{lc}
            \toprule
            Approach & F1 Score \\
            \midrule
            Small model only & 58.2\% \\
            Large model only & 87.5\% \\
            Uncertainty-based system~\citep{narayan2022predicting} & 74.7\% \\
            Ours & 92.1\% \\
            \bottomrule
        \end{tabular}
        \captionof{table}{F1 Scores of different approaches}
        \label{tab:main_result}
    \end{minipage}
\end{figure}
\end{comment}

\begin{table}[]
    \centering
    \begin{tabular}{lcccc}
        \toprule
        Approach & F1 (\%) & Large Model (\%)~\tablefootnote{Large Model (\%) refers to the percentage of test data instances that were processed using the large model.} & Time (s) & Energy (kJ)\\
        \midrule
        Small model only~\citep{kim2021vilt} & 58.2 & 0 & 18.25 & 2.32\\
        Large model only~\citep{li2024llava} & 87.5 & 100 & 663.1 & 190.73\\
        Uncertainty-based system~\citep{narayan2022predicting} & 76.1 & 60 & 386.91 & 113.04 \\
        Ours & \textbf{92.1} & 60 & 405.16 & 115.36 \\
        \bottomrule
    \end{tabular}
    \captionof{table}{Results of different approaches. }
    \label{tab:main_result}
\end{table}

\paragraph{Computational Cost Evaluation}
We also evaluated the computational cost of our model in terms of energy consumption and runtime compared to using a large model only solution, following the method described in Appendix~\ref{appendix:cost}. As illustrated in Figure \ref{fig:energy}, our switcher model solution, which defers judgment to the large model 60\% of the time (corresponding to the switcher model's behavior on the test data), achieved a significant reduction in energy consumption. Specifically, the energy used dropped from 0.0530 kWh to 0.0320 kWh, a 39.5\% reduction in energy usage. Also, as shown in Figure~\ref{fig:runtime}, the time taken for processing decreased from 663.1 seconds to 405.16 seconds, a 38.9\% reduction in time taken. Thus, our system not only improves detection quality but also enhances detection latency, energy efficiency, and overall computational cost.

\section{Conclusion and Future Works}
In this paper, we introduced a novel Dual-Model Distillation (DMD) method to develop a lightweight hybrid edge-cloud solution that efficiently balances computational cost and accuracy, enabling deployment on resource-constrained edge devices with support from a large model in the cloud. Although our framework is designed for long video inputs, our current experiments are limited to single-frame video inputs. In future work, we plan to transition to video-based inputs and test our method on more diverse action classification datasets. Overall, we expect our framework to be adaptable and applicable to other real-world applications beyond fall detection. Moreover, we anticipate that DMD will advance the research frontier of knowledge alignment across multiple models in a variety of domains.

\begin{ack}
    We would like to thank Evan Li and Ryan Tellado for their insightful feedback and contributions to the development of this paper.
\end{ack}

% In conclusion, our proposed system introduces an efficient dual-model framework that balances computational cost with high accuracy and has higher efficacy than the smaller model or larger model. By leveraging a switcher model, we defer complex cases to a larger model, optimizing resource usage without sacrificing performance. The novel dual-model distillation technique allows us to train the switcher using both models' knowledge, eliminating the need for labeled data. Our results demonstrate the system's ability to enhance prediction accuracy while minimizing computational overhead. Future work will focus on extending the system to video inputs, improving its effectiveness in dynamic, real-world scenarios.

\bibliographystyle{plainnat} % We choose the "plain" reference style
\bibliography{sample} % Entries are in the refs.bib file

\begin{thebibliography}{19}
\providecommand{\natexlab}[1]{#1}
\providecommand{\url}[1]{\texttt{#1}}
\expandafter\ifx\csname urlstyle\endcsname\relax
  \providecommand{\doi}[1]{doi: #1}\else
  \providecommand{\doi}{doi: \begingroup \urlstyle{rm}\Url}\fi

\bibitem[Chen and Ran(2019)]{chen2019deep}
Jiasi Chen and Xukan Ran.
\newblock Deep learning with edge computing: A review.
\newblock \emph{Proceedings of the IEEE}, 107\penalty0 (8):\penalty0 1655--1674, 2019.

\bibitem[Chung et~al.(2024)Chung, Hou, Longpre, Zoph, Tay, Fedus, Li, Wang, Dehghani, Brahma, et~al.]{chung2024scaling}
Hyung~Won Chung, Le~Hou, Shayne Longpre, Barret Zoph, Yi~Tay, William Fedus, Yunxuan Li, Xuezhi Wang, Mostafa Dehghani, Siddhartha Brahma, et~al.
\newblock Scaling instruction-finetuned language models.
\newblock \emph{Journal of Machine Learning Research}, 25\penalty0 (70):\penalty0 1--53, 2024.

\bibitem[Dubey et~al.(2024)Dubey, Jauhri, Pandey, Kadian, Al-Dahle, Letman, Mathur, Schelten, Yang, Fan, et~al.]{dubey2024llama}
Abhimanyu Dubey, Abhinav Jauhri, Abhinav Pandey, Abhishek Kadian, Ahmad Al-Dahle, Aiesha Letman, Akhil Mathur, Alan Schelten, Amy Yang, Angela Fan, et~al.
\newblock The llama 3 herd of models.
\newblock \emph{arXiv preprint arXiv:2407.21783}, 2024.

\bibitem[Elwaly and Abdellatif(2023)]{Elwaly2023dataset}
Ahmad Elwaly and Ahmed Abdellatif.
\newblock Fall detection - eldercare robot, February 2023.
\newblock URL \url{https://www.kaggle.com/datasets/elwalyahmad/fall-detection}.

\bibitem[Elwaly et~al.(2024)Elwaly, Abdellatif, and El-Shaer]{elwaly2024new}
Ahmad Elwaly, A~Abdellatif, and Y~El-Shaer.
\newblock New eldercare robot with path-planning and fall-detection capabilities.
\newblock \emph{Applied Sciences}, 14\penalty0 (6):\penalty0 2374, 2024.

\bibitem[Gou et~al.(2021)Gou, Yu, Maybank, and Tao]{gou2021knowledge}
Jianping Gou, Baosheng Yu, Stephen~J Maybank, and Dacheng Tao.
\newblock Knowledge distillation: A survey.
\newblock \emph{International Journal of Computer Vision}, 129\penalty0 (6):\penalty0 1789--1819, 2021.

\bibitem[Hinton(2015)]{hinton2015distilling}
Geoffrey Hinton.
\newblock Distilling the knowledge in a neural network.
\newblock \emph{arXiv preprint arXiv:1503.02531}, 2015.

\bibitem[Jiang et~al.(2023)Jiang, Sablayrolles, Mensch, Bamford, Chaplot, Casas, Bressand, Lengyel, Lample, Saulnier, et~al.]{jiang2023mistral}
Albert~Q Jiang, Alexandre Sablayrolles, Arthur Mensch, Chris Bamford, Devendra~Singh Chaplot, Diego de~las Casas, Florian Bressand, Gianna Lengyel, Guillaume Lample, Lucile Saulnier, et~al.
\newblock Mistral 7b.
\newblock \emph{arXiv preprint arXiv:2310.06825}, 2023.

\bibitem[Kim et~al.(2021)Kim, Son, and Kim]{kim2021vilt}
Wonjae Kim, Bokyung Son, and Ildoo Kim.
\newblock Vilt: Vision-and-language transformer without convolution or region supervision, 2021.

\bibitem[Li et~al.(2024)Li, Zhang, Zhang, Zhang, Li, Li, Ma, and Li]{li2024llava}
Feng Li, Renrui Zhang, Hao Zhang, Yuanhan Zhang, Bo~Li, Wei Li, Zejun Ma, and Chunyuan Li.
\newblock Llava-next-interleave: Tackling multi-image, video, and 3d in large multimodal models.
\newblock \emph{arXiv preprint arXiv:2407.07895}, 2024.

\bibitem[Lin et~al.(2023)Lin, Zhu, Ye, Ning, Jin, and Yuan]{lin2023video}
Bin Lin, Bin Zhu, Yang Ye, Munan Ning, Peng Jin, and Li~Yuan.
\newblock Video-llava: Learning united visual representation by alignment before projection.
\newblock \emph{arXiv preprint arXiv:2311.10122}, 2023.

\bibitem[Maldonado-Bascon et~al.(2019)Maldonado-Bascon, Iglesias-Iglesias, Mart{\'\i}n-Mart{\'\i}n, and Lafuente-Arroyo]{maldonado2019fallen}
Saturnino Maldonado-Bascon, Cristian Iglesias-Iglesias, Pilar Mart{\'\i}n-Mart{\'\i}n, and Sergio Lafuente-Arroyo.
\newblock Fallen people detection capabilities using assistive robot.
\newblock \emph{Electronics}, 8\penalty0 (9):\penalty0 915, 2019.

\bibitem[Minaee et~al.(2024)Minaee, Mikolov, Nikzad, Chenaghlu, Socher, Amatriain, and Gao]{minaee2024large}
Shervin Minaee, Tomas Mikolov, Narjes Nikzad, Meysam Chenaghlu, Richard Socher, Xavier Amatriain, and Jianfeng Gao.
\newblock Large language models: A survey.
\newblock \emph{arXiv preprint arXiv:2402.06196}, 2024.

\bibitem[Narayan et~al.(2022)Narayan, Jiang, Zhao, and Kumar]{narayan2022predicting}
Taman Narayan, Heinrich Jiang, Sen Zhao, and Sanjiv Kumar.
\newblock Predicting on the edge: Identifying where a larger model does better.
\newblock \emph{arXiv preprint arXiv:2202.07652}, 2022.

\bibitem[Sawik et~al.(2023)Sawik, Tobis, Baum, Suwalska, Kropi{\'n}ska, Stachnik, P{\'e}rez-Bernabeu, Cildoz, Agustin, and Wieczorowska-Tobis]{sawik2023robots}
Bartosz Sawik, S{\l}awomir Tobis, Ewa Baum, Aleksandra Suwalska, Sylwia Kropi{\'n}ska, Katarzyna Stachnik, Elena P{\'e}rez-Bernabeu, Marta Cildoz, Alba Agustin, and Katarzyna Wieczorowska-Tobis.
\newblock Robots for elderly care: Review, multi-criteria optimization model and qualitative case study.
\newblock In \emph{Healthcare}, volume~11, page 1286. MDPI, 2023.

\bibitem[Wang et~al.(2023)Wang, Bochkovskiy, and Liao]{wang2023yolov7}
Chien-Yao Wang, Alexey Bochkovskiy, and Hong-Yuan~Mark Liao.
\newblock Yolov7: Trainable bag-of-freebies sets new state-of-the-art for real-time object detectors.
\newblock In \emph{Proceedings of the IEEE/CVF conference on computer vision and pattern recognition}, pages 7464--7475, 2023.

\bibitem[Wang and Zhang(2011)]{wang2011utilizing}
Jian Wang and Yi~Zhang.
\newblock Utilizing marginal net utility for recommendation in e-commerce.
\newblock In \emph{Proceedings of the 34th international ACM SIGIR conference on Research and development in Information Retrieval}, pages 1003--1012, 2011.

\bibitem[Wu et~al.(2023)Wu, Fei, Qu, Ji, and Chua]{wunext}
Shengqiong Wu, Hao Fei, Leigang Qu, Wei Ji, and Tat-Seng Chua.
\newblock Next-gpt: Any-to-any multimodal llm.
\newblock In \emph{Forty-first International Conference on Machine Learning}, 2023.

\bibitem[Zhao et~al.(2023)Zhao, Zhou, Li, Tang, Wang, Hou, Min, Zhang, Zhang, Dong, et~al.]{zhao2023survey}
Wayne~Xin Zhao, Kun Zhou, Junyi Li, Tianyi Tang, Xiaolei Wang, Yupeng Hou, Yingqian Min, Beichen Zhang, Junjie Zhang, Zican Dong, et~al.
\newblock A survey of large language models.
\newblock \emph{arXiv preprint arXiv:2303.18223}, 2023.

\end{thebibliography}
%%%%% This workshop doesn't require NeurIPS submission checklist
%%%%% The submission guideline is moved to neurips_submission_checklist file

%%%%%%%%%%%%%%%%%%%%%%%%%%%%%%%%%%%%%%%%%%%%%%%%%%%%%%%%%%%%

\appendix

\section{Appendix and Supplemental Materials}

% \subsection{Acknowledgements}

% We would like to gratefully thank Diji Yang and Lei Ding for assisting us with their feedback and thoughts. We also would like to thank Evan Li and Ryan Tellado for their insightful ideas and contributions. 

%TODO Overview in a condensed paragraph (clearly describe the process of doing it (need to explain table))
% \subsection{Overview}

\subsection{Datasets}
\label{appendix:datasets}
We used the Fall Detection - Eldercare Robot dataset for training, validation and testing. It includes 1061 images, each having a data entry in YOLO v7 PyTorch format \cite{wang2023yolov7}. The data entry includes a binary fall/no-fall classification label, the coordinates of the center of the bounding box containing the person, and the width and height of the bounding box. The creators of the dataset oriented pixel data automatically using EXIF-orientation stripping and stretched to resize it to 640x640. When exported, each image had a 50\% probability of being laterally inverted and a one-in-three chance of undergoing one of the following operations: a 90-degree clockwise rotation, a 90-degree anti-clockwise rotation, or none at all. Of the 1061 images, we follow the commonly used split, i.e., 742 for training, 212 for validation, and 106 for testing.

\subsection{Data Generation and Formatting using Dual-Model Distillation}
In order to prepare and standardize our data for training, we implemented the DMD process as described in this section. As discussed in Section~\ref{sec:approach}, the only information we needed to train the switcher model is the nodes of the last hidden layer for each given input and whether the large and small models agree, referred to as the alignment between the models. We used our novel DMD framework to refine these data for training: by inputting each image into both models with a prompt telling them to report if a fall occurred, we extracted their outputs and recorded whether the models agreed. The values of the nodes in the last hidden layer of ViLT were extracted and saved accordingly. Ultimately, our refined data consisted of only three fields, making the DMD process compatible with any image or video dataset. An example of the formatted data after the DMD process is given as following. 
\begin{verbatim}
[
    {
        "image_path": "path_to_image/img.jpg",
        "last_hidden_layer": [
            1.369998574256897,
            ...
        ],
        "label": 1
    },...
]
\end{verbatim}

\subsection{Implementation Details}
\label{appendix:implementation_details}
% goes to appendix
\paragraph{Model Design} To make our switcher model lightweight yet effective, we implemented a Multi-Layer Perceptron (MLP) with input size 1536, two hidden layers of size 512 and 128, and an output layer of size 1. Due to this structure, it can be easily implemented in a local system with little additional cost or memory. However, we also found that by using the hidden representation of the last hidden layer from the small model, the MLP demonstrates the capability of understanding the reasoning of the small model, therefore making better alignment predictions.
%%%%% training detail of the swither model
% As for training the switcher model, we made a lightweight Multi-Layer Perceptron (MLP) Model capable of representing key components in the small model's reasoning. Using an MLP ensures that the switcher model can decide how likely both models will agree while keeping the computational cost to a minimum. 

\paragraph{Training Details} As for the training settings, we used a learning rate of 0.0001 and a dropout rate of 0.3. Using the refined data mentioned above, we calculated the Binary Cross-Entropy loss (BCELoss) followed by the Logits function (which applies the BCELoss function after the Sigmoid function), and updated weights through backpropagation. The training early stopped at 100 epochs, where the model achieved an F1 score of 79.2\% and an accuracy of 82.2\% stably. 

\subsection{Threshold for Offload Decision} 
\label{appendix:threshold}
% to be updated
To determine the optimal threshold for deferring judgment to the large model at inference time, we analyzed the switcher model's performance on the training data. 

\begin{figure}[htbp]
    \centering
    \begin{minipage}{0.45\linewidth}
        \centering
        \includegraphics[width=\linewidth]{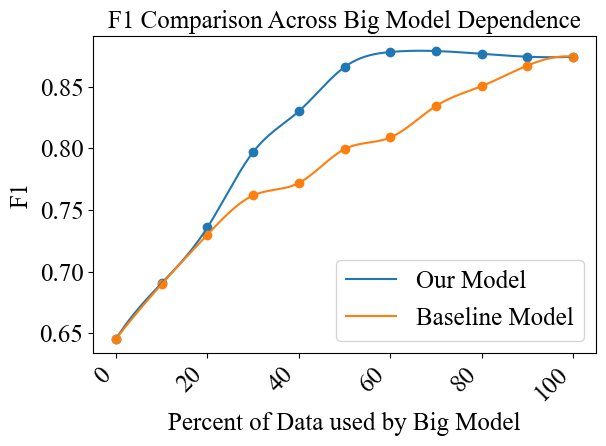}
        \caption{Combined and baseline model accuracy against the percentage of data deferred to the large model in the training data split}
        \label{fig:F1_data_percentage_plot_on_trainset}
    \end{minipage}%
    \hfill
    \begin{minipage}{0.45\linewidth}
        \centering
        \includegraphics[width=\linewidth]{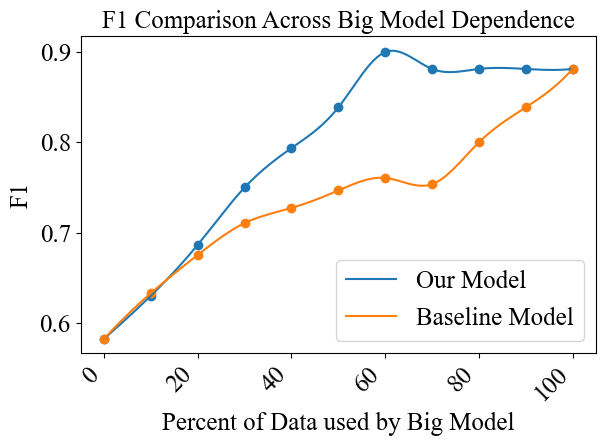}
        \caption{Combined and baseline model accuracy against the percentage of data deferred to the large model in the testing data split}
        \label{fig:F1_data_percentage_plot_on_testset}
    \end{minipage}
\end{figure}

We collected the results through our switcher model after all the training data was processed. We sorted these results in ascending order of alignment probability and split the results into 10 "buckets". Using this "buckets" strategy on the training set, we then plotted the graph of the combined model F1 score against the percentage of data deferred to the large model. For comparison purposes, we also plot the baseline (a standard uncertainty-based switcher model) in Figure\ref{fig:F1_data_percentage_plot_on_trainset}.

By observing the training curve of the combined model, we chose the optimal threshold value 60\%, shown as the peak of the curve in Figure \ref{fig:F1_data_percentage_plot_on_trainset}, for testing. For the baseline, uncertainty-based system, we also used 60\%, because the baseline uncertainty doesn't have a clear peak to indicate how much data should be offloaded to the large model. Keeping this threshold consistent helps us compare performance fairly. The corresponding results are shown in Table~\ref{tab:main_result}.

To explore how the threshold value influences the final testing performance, we further applied our "buckets" strategy to test data for both methods and analyzed the results. The corresponding results are shown in Figure \ref{fig:F1_data_percentage_plot_on_testset}. As a note, this analysis of accuracy against percent of data deferred was used extensively in \citep{narayan2022predicting} and can be extended to any framework involving a switcher model.

% this table will be changed

% \subsection{Results Graphs}

\begin{figure}[htbp]
    \centering
    \begin{minipage}{0.48\linewidth}
        \centering
        \includegraphics[width=\linewidth]{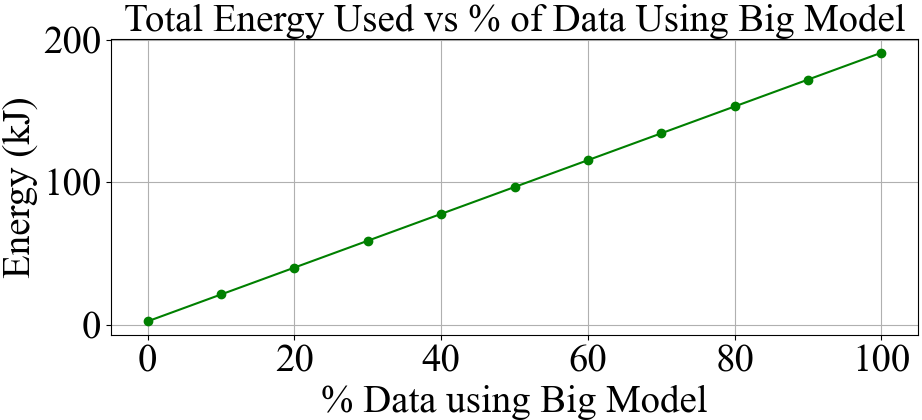}
        \caption{Total energy consumed against the percentage of data deferred to big model}
        \label{fig:energy}
    \end{minipage}%
    \hfill
    \begin{minipage}{0.48\linewidth}
        \centering
        \includegraphics[width=\linewidth]{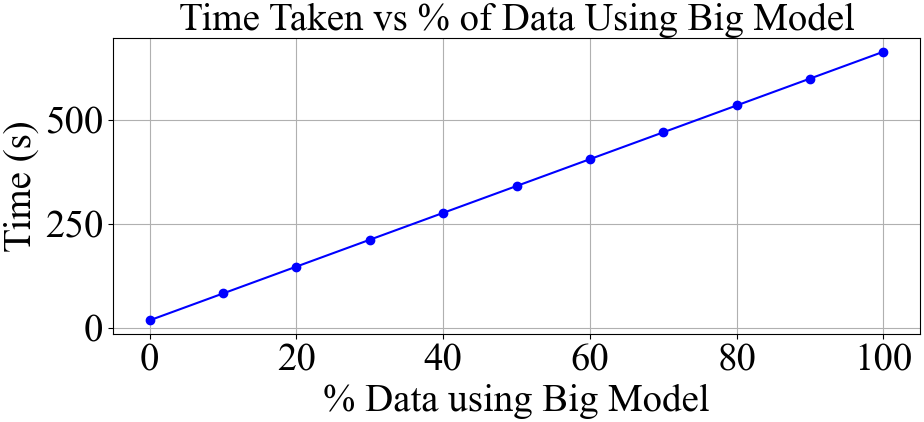}
        \caption{Runtime of combined model against percentage of data deferred to big model}
        \label{fig:runtime}
    \end{minipage}
\end{figure}

\begin{comment}
\begin{figure}[htbp]
    \centering
    \begin{minipage}{0.45\linewidth}
        \centering
        \includegraphics[width=\linewidth]{train_results.png}
        \caption{Combined and baseline model accuracy against the percentage of data deferred to the large model in the training data split}
        \label{fig:fig2}
    \end{minipage}%
    \hfill
    \begin{minipage}{0.45\linewidth}
        \centering
        \begin{tabular}{lc}
            \toprule
            Approach & F1 Score \\
            \midrule
            Small model only & 64.5\% \\
            Large model only & 87.4\% \\
            Uncertainty-based system~\citep{narayan2022predicting} & 79.9\% \\
            Ours & 86.6\% \\
            \bottomrule
        \end{tabular}
        \captionof{table}{F1 Scores of different approaches}
        \label{tab:main_result}
    \end{minipage}
\end{figure}
\end{comment}

\subsection{Computational Cost Evaluation}
\label{appendix:cost}
Finally, we compared our model's computational cost to our baselines. We measured the total energy consumption of our system using NVIDIA’s \texttt{nvidia-smi} tool and tracked the time taken for processing with Python’s \texttt{time} library on our training dataset. To explore the tradeoff in computational cost for performance, we graphed both energy consumption and time against the percentage of test cases where the large model was involved. Our simplistic analysis can be applied to other metrics of computational cost and to analysis of other models. The results are illustrated in Figure~\ref{fig:energy} and Figure~\ref{fig:runtime}.

\end{document}